\documentclass[11pt]{article}
\PassOptionsToPackage{table}{xcolor}

\usepackage{listings}
\usepackage{xcolor}

\usepackage[final]{acl}

\usepackage{times}
\usepackage{latexsym}
\usepackage[T1]{fontenc}

\usepackage[utf8]{inputenc}

\usepackage{microtype}
\usepackage{todonotes}
\usepackage{inconsolata}
\usepackage{amsmath,amsfonts}
\usepackage{algorithmic}
\usepackage{textcomp}
\usepackage{caption}
\usepackage{subcaption}
\usepackage{multirow}
\usepackage{graphicx}
\usepackage{multicol}
\usepackage[titletoc]{appendix}
\usepackage{booktabs}
\usepackage{enumitem}
\usepackage{listings}
\usepackage{breakcites}
\usepackage{algorithm} 
\usepackage{algorithmic}
\usepackage{ulem}
\usepackage{wrapfig}

\usepackage{lineno}
\def\gobble#1{}
\usepackage{xparse}
\usepackage[most]{tcolorbox}
\usepackage{pgf}
\usepackage{collcell}

\newcommand{\ApplyColorIfNeeded}[1]{%
    \pgfmathparse{#1}%
    \ifnum\pgfmathresult>0
        \pgfmathsetmacro{\CellColorIntensity}{(\pgfmathresult ) * 10} 
        \edef\temp{\noexpand\cellcolor{black!\CellColorIntensity}}\temp #1
    \else
        #1
    \fi
}

\newtcolorbox[auto counter]{mybox}[2][]{
  colback=bg, 
  colframe=blue!75!black,
  fonttitle=\bfseries,
  coltitle=white,
  colbacktitle=blue!85!black,
  title={Box \thetcbcounter: #2},
  label={box@\thetcbcounter},
  sharp corners,
  halign title=center, 
  width=0.48\columnwidth, 
  #1,
}

\colorlet{bg}{white!90!blue}
\usepackage{mdframed}

\usepackage{etoolbox}
\title{Towards Effectively Leveraging Execution Traces \\for Program Repair with Code LLMs}

\author{Mirazul Haque\textsuperscript{1}\footnotemark[1] , Petr Babkin\textsuperscript{2}\footnotemark[1] , Farima Farmahinifarahani\textsuperscript{2}, Manuela Veloso\textsuperscript{1} \\
  J. P. Morgan AI Research, \{\textsuperscript{1}New York, \textsuperscript{2}Palo Alto\}\\
  \texttt{\{first.last\}@jpmchase.com}}

\begin{document}
\maketitle
\begin{abstract}
Large Language Models (LLMs) show promising performance on various programming tasks, including Automatic Program Repair (APR).
However, most approaches to LLM-based APR are limited to the static analysis of the programs, while disregarding their runtime behavior.
Inspired by knowledge-augmented NLP, in this work, we aim to remedy this potential blind spot by augmenting standard APR prompts with program execution traces.
We evaluate our approach using the GPT family of models on three popular APR datasets. 
Our findings suggest that simply incorporating execution traces into the prompt provides a limited performance improvement over trace-free baselines, in only 2 out of 6 tested dataset / model configurations. 
We further find that the effectiveness of execution traces for APR diminishes as their complexity increases. 
We explore several strategies for leveraging traces in prompts
and demonstrate that LLM-optimized prompts help outperform trace-free prompts more consistently.
Additionally, we show trace-based prompting to be superior to finetuning a smaller LLM on a small-scale dataset; and conduct probing studies reinforcing the notion that execution traces can complement the reasoning abilities of the LLMs.

\end{abstract}
\renewcommand{\thefootnote}{\fnsymbol{footnote}}
\footnotetext[1]{equal contribution.}
\renewcommand{\thefootnote}{\arabic{footnote}}

\section{Introduction}
Automatic Program Repair (APR) is a critical challenge in software engineering, aiming to reduce human effort in debugging and fixing software defects. Software bugs can lead to significant security vulnerabilities, financial losses, and system failures, necessitating efficient repair mechanisms. While large language models (LLMs) have demonstrated remarkable capabilities in generating and modifying code, their effectiveness in APR remains constrained by their reliance on static code analysis. Debugging complex software issues often necessitates a deeper understanding of the program’s execution behavior, including variable modifications and control flow changes, which conventional Deep Learning-based and LLM-based APR approaches fail to capture effectively ~\cite{xia2022less,jiang2023impact, tian2023chatgpt, invariants52366}.

\begin{figure}[t]
\begin{lstlisting}[language=Python]
### Buggy Program:
def search(x, seq):
    index = 0
    def helper(index):
        if not seq:
            return 0
        elif x <= seq[index]:
            return index
        else:
            if index + 1 >= len(seq):
                return index + 1
            else:
                return helper(index+1)

### Failing test case:
result = search(42, (-5, 1, 3, 5, 7, 10))
assert result == 6,                    'Expected 6 but got %s' % result
AssertionError: Expected 6 but got None

### Execution trace:
Starting var:.. x = 42
Starting var:.. seq = (-5, 1, 3, 5, 7, 10)
call        10 def search(x, seq):
line        11     index = 0
New var:....... index = 0
line        12     def helper(index):
New var:....... helper = <function search.<locals> .helper at 0x7fd455b89040>
return      12     def helper(index):
Return value:.. None
\end{lstlisting}
\label{fig:trace-example}
\caption{Example buggy program, a failing test case and its execution trace. While the failure message simply indicates the output is wrong, the execution trace provides a detailed explanation how it was produced.}
\end{figure}
Recent advancements in knowledge-augmented NLP have emphasized integrating external information into language models to enhance reasoning and accuracy. Inspired by this, our research explores augmenting LLM-based automated program repair (APR) with program execution traces—structured runtime data that reveal a program's actual behavior. These traces provide diagnostic insights beyond static code analysis. By embedding them into repair prompts, we aim to bridge the gap between static and dynamic program understanding, aligning with trends in knowledge-augmented NLP that leverage external sources to enhance language model capabilities.

We frame our work in terms of three research questions (RQs). 
In RQ1 (Section \ref{sec:RQ1}), our objective is to quantify the gains from incorporating execution traces into the APR prompt over the prompts only containing the failing test case as well as the 
trace-free chain-of-thought prompting baseline~\cite{chen2023teaching}.
We find that simply adding the execution trace does not consistently outperform trace-free prompts.



To inform a more finegrained approach, in RQ2 (Section \ref{sec:RQ2}), we analyze the relationship between trace complexity and the likelihood of the LLM producing a working fix.
To measure this complexity, we consider two parameters: trace length and the number of variable modifications. 
We find that the effectiveness of trace-based prompts decreases with the growing length and number of variable assignments.


Motivated by this finding, in RQ3 (Section \ref{sec:RQ3}), we aim at improving the consistency of trace-based APR by experimenting with three different representations of execution traces: traces in a collated format, LLM-optimized traces, and a trace representation conditionally selected based on querying the LLM's confidence. 
We find that LLM-optimized trace-based prompts provide the most consistent results with respect to program repair. 

We additionally perform two follow-up studies: in the first, we compare our trace-based prompting approach with a fine-tuned baseline inspired by TraceFixer \cite{bouzenia2023tracefixer}; and in the second one, we directly probe the LLM on two trace understanding tasks.




The rest of the paper is organized as follows. In Section \ref{sec:back}, we discuss the related work and how it differs from our approach.
Section \ref{sec:rq12} details our methodological setup and covers RQ1 and RQ2. 
Section \ref{sec:rq3} covers RQ2 and in Section \ref{sec:disc}, we discuss the additional studies.





\section{Related Work}
\label{sec:back}

Recent work looked into augmenting code LLMs with execution information to improve performance on downstream tasks, including APR.

SelfAPR \cite{ye2022selfapr} proposed to use compiler and test diagnostics during self-supervised training of the language model for improving APR.  
Additionally, several works have proposed the use of execution traces for pretraining code LLMs. 
In TRACED ~\cite{ding2023traced}, authors finetuned a BERT-like model to predict execution paths and quantized values, which allowed it to outperform an AST-based UniXcoder~\cite{guo2022unixcoderunifiedcrossmodalpretraining} on clone detection and vulnerability detection.
Whereas, Liu et al.’s program state prediction pre-training improved code search and generation~\cite{liu2023code}.
Finally, TraceFixer, based on CodeT5 and finetuned with execution traces, showed a 13\% improvement in APR on synthetic bugs over the code-only baseline but struggled with real bugs, hinting at potential generalization limitations~\cite{bouzenia2023tracefixer}. 

Among training-free approaches, 
Self-Debug \cite{chen2023teaching} improved program generation by generating code explanations directly from the LLM, in a chain-of-thought fashion, as part of solving the APR task.

To the best of our knowledge, all of these works do not consider the effect of putting execution traces in the prompt of a pretrained LLM. 

\section{Analyzing the Impact of Execution Traces on Program Repair}
\label{sec:rq12}

In this section, we analyze the effects of adding traces in the LLM prompt on APR performance, compared to two trace-free baselines ~\cite{ye2022selfapr}, ~\cite{chen2023teaching}.
Additionally, we perform a differentiated analysis of APR performance based on trace complexity.
We formulate the corresponding two research questions as follows.

\noindent\textbf{RQ1.} Are prompts with execution traces more effective at program repair than prompts without traces?

\noindent\textbf{RQ2.} How does trace complexity affect the effectiveness of trace-based prompts?

\subsection{Set Up}
\noindent\textbf{Datasets.} 
We surveyed 15 popular datasets across Python, Java, C++, and other major languages, focusing on dataset size, program diversity, unit test availability, and dataset origin (e.g., self-contained algorithmic problems like CodeNet \cite{puri2021codenet} or full open-source projects like PyTraceBugs \cite{akimova2021pytracebugs}). While realistic datasets are ideal, evaluating them requires significant manual effort due to complex dependencies. Algorithmic datasets offer advantages like manageable length and easily testable, self-contained functions, enabling trace generation through execution.

We selected three APR datasets: Refactory \cite{hu2019re}, RunBugRun \cite{prenner2023runbugrun}, and HumanEval-Java \cite{jiang2023impact}. Refactory includes nearly 2000 faulty Python programs submitted by students, enabling coverage of diverse mistakes. RunBugRun, derived from CodeNet, contains a quarter million submissions for 4000 distinct problems; we sampled 1000 Python bugs for evaluation. HumanEval, originally for Python, was adapted into HumanEval-Java, injecting synthetic bugs for APR testing.

Each dataset includes at least 5 test cases per problem. For RunBugRun, we implemented a wrapper to handle input/output via standard input and print statements for accurate result comparison.

\noindent\textbf{Models.} With the landscape of state-of-the-art code LLMs rapidly changing, we chose use two most widely studied commercial models from OpenAI for ease of comparison with other work: GPT-3.5 Turbo ~\cite{ouyang2022training} and GPT-4~\cite{OpenAI2023GPT4TR}. 
These two models represent two different performance tiers both in terms of the number of parameters and different release timelines, hence, studying these models could shine the light on the LLMs' evolving ability to reason about program execution across product generations. While there is undoubtedly scope for including more proprietary as well as open source models, given our narrow focus on traces, we leave this to be explored in future work.

\noindent\textbf{Execution Traces Generation.} 
As the program is being executed, it is possible to step through it programmatically, while also capturing every change to the function's variables, akin to interactive debugging. PySnooper~\cite{rachum2019pysnooper} library for Python provides this functionality via a decorator that can be added to a function of interest to automatically log state changes, such as variable initialization and modification, subroutine calls, returned values, and runtime exceptions. Crucially, each state change reference a specific line of code on which it occurred. Before appending traces to the prompt we perform basic postprocessing, including the removal of timestamps and stripping of terminal formatting command sequences.

\noindent\textbf{Prompt Types.} 
We follow the instruction template for complete function generation used by Xia et al. (\citeyear{ICSE10172803}), expanding it with two additional types of information, namely, a failing test case (henceforth, referred to as \textit{Error Prompts}) and a program execution trace (referred as \textit{Trace Prompts}). 
We offer our rationale for these choices, along with other prompt types considered, in Appendix \ref{sec:prompt-rationale}.

To ensure the prompt and response fit within the GPT model context size, we truncate the content of the prompt if the number of lines exceeds 200.

\begin{table*}[t]
  \centering
  \resizebox{0.85\textwidth}{!}{
    \begin{tabular}{c|cl|r|r|r|r|r|r}
    \toprule
    \textbf{Model} & \textbf{Dataset} & \textbf{Method} & \textbf{\# FPs} & \textbf{\# Fixes} & \textbf{\# CF} & \textbf{\# CP} & \textbf{CFA} & \textbf{CPA} \\
    \midrule
    \multirow{9}[1]{*}{\textbf{GPT-3.5}} 
      & \multirow{3}[1]{*}{\textbf{Refactory}} & \textbf{Self-Debug} & \multirow{3}[1]{*}{138} & \multirow{3}[1]{*}{579} & 244 & 73 & 0.421 & 0.529 \\
      & & \textbf{Error Prompt} & & & \textbf{304} & \textbf{91} & \textbf{0.525} & \textbf{0.659} \\
      & & \textbf{Trace Prompt} & & & 295 & 87 & 0.509 & 0.630 \\
    \cline{2-9}
      & \multirow{3}[1]{*}{\textbf{HumanEval-Java}} & \textbf{Self-Debug} & \multirow{3}[1]{*}{157} & \multirow{3}[1]{*}{634} & 210 & 75 & 0.331 & 0.477 \\
      & & \textbf{Error Prompt} & & & \textbf{241} & 85 & \textbf{0.380} & 0.541 \\
      & & \textbf{Trace Prompt} & & & 212 & \textbf{86} & 0.334 & \textbf{0.547} \\
    \cline{2-9}
      & \multirow{3}[1]{*}{\textbf{RunBugRun}} & \textbf{Self-Debug} & \multirow{3}[1]{*}{456} & \multirow{3}[1]{*}{559} & 151 & 132 & 0.270 & 0.289 \\
      & & \textbf{Error Prompt} & & & \textbf{260} & \textbf{221} & \textbf{0.465} & \textbf{0.484} \\
      & & \textbf{Trace Prompt} & & & 249 & 216 & 0.445 & 0.473 \\
    \midrule
    \multirow{9}[1]{*}{\textbf{GPT-4}} 
      & \multirow{3}[1]{*}{\textbf{Refactory}} & \textbf{Self-Debug} & \multirow{3}[1]{*}{138} & \multirow{3}[1]{*}{579} & 414 & 117 & 0.715 & 0.847 \\
      & & \textbf{Error Prompt} & & & \textbf{458} & \textbf{122} & \textbf{0.791} & \textbf{0.884} \\
      & & \textbf{Trace Prompt} & & & 427 & 113 & 0.737 & 0.818 \\
    \cline{2-9}
      & \multirow{3}[1]{*}{\textbf{HumanEval-Java}} & \textbf{Self-Debug} & \multirow{3}[1]{*}{157} & \multirow{3}[1]{*}{634} & 312 & 105 & 0.492 & 0.668 \\
      & & \textbf{Error Prompt} & & & 313 & 104 & 0.493 & 0.662 \\
      & & \textbf{Trace Prompt} & & & \textbf{324} & \textbf{112} & \textbf{0.511} & \textbf{0.713} \\
    \cline{2-9}
      & \multirow{3}[1]{*}{\textbf{RunBugRun}} & \textbf{Self-Debug} & \multirow{3}[1]{*}{456} & \multirow{3}[1]{*}{559} & \textbf{337} & \textbf{287} & \textbf{0.602} & \textbf{0.629} \\
      & & \textbf{Error Prompt} & & & 296 & 264 & 0.529 & 0.578 \\
      & & \textbf{Trace Prompt} & & & \textit{312} & \textit{266} & \textit{0.558} & \textit{0.583} \\
    \bottomrule
    \end{tabular}
  }
    \caption{\textbf{RQ1 Quantitative Results.} FP = Faulty Programs, CF = Correct Fixes, CP = Correct Programs, \\CFA = Correct Fix Accuracy, CPA = Correct Program Accuracy.}
  \label{tab:Per_c10_2_2}
\end{table*}

\noindent\textbf{Baseline.} 
We consider the Prompt-based baseline Self-Debug.
With this baseline, we explore prompting LLMs using execution traces generated by LLMs themselves (instead of actual program execution traces). This baseline inspired by Self-Debugging~\cite{chen2023teaching} where LLMs are prompted to debug their own generated code. In particular, we draw inspirations from the \textit{Explanation} step of this work where the model is asked to generate execution traces for a predicted code. We tailored Self-Debugging's prompts to fit our usecase: in our prompts, we provide LLMs with a program and a test case feedback, and ask them to trace through the execution of the program and determine the needed fix, and correct the function accordingly.  We perform these experiments with both GPT-3.5 and GPT-4.



\noindent\textbf{Metrics.} 
In previous work on APR, models are evaluated either at the granularity of distinct bugs solved~\cite{ICSE10172803,jiang2023impact} or individual test cases passed \cite{tian2023chatgpt}. In contrast, we generate multiple prompts for each program tailored to a specific failing test case and its corresponding execution trace. Rather than aggregating predictions from multiple samples, we generate a single prediction per test case-specific prompt and aggregate across prompts when computing metrics. The key metrics are \textit{\textbf{Correct Fix Accuracy (CFA)}}, the percentage of fixes passing all test cases, and \textit{\textbf{Correct Program Accuracy (CPA)}}, the percentage of programs with at least one correct fix. We do not report test case-level accuracy, as it can be too lenient and doesn't account for variations in the number of test cases per program.
\subsection{RQ1. Are prompts with execution traces more effective at program repair than prompts without traces?}
\label{sec:RQ1}
In this research question, our objective is to evaluate the effectiveness of including program execution traces into LLM prompts, for solving APR tasks, compared to the baselines. 
The effectiveness is measured through reporting CPA and CFA. 
%
%
%
%
%
The evaluation results can be found in Table \ref{tab:Per_c10_2_2}. The number of faulty programs, number of fixes, and total test cases are the same for all types of prompts per each dataset.  

Across the board, the Self-Debug baseline performs the worst except in one configuration using GPT-4 on the RunBugRun dataset. This general outcome is unsurprising as having the LLM generate an execution trace could introduce hallucination and thus undermine the resulting fixes.
For both GPT-3.5 and GPT-4 on the Refactory dataset, prompts including just a failing test case decisively outperform ones with execution traces by multiple percentage points on both fix accuracy and program accuracy. 

On the HumanEval-Java dataset with GPT-3.5, error-only prompts are only ahead of trace-based prompts in terms of fix accuracy but are slightly behind in program accuracy. Meanwhile, with GPT-4, trace-based prompts consistently outperform error-only prompts on both metrics. 
Results on RunBugRun paint a similar picture, where GPT-3.5 doesn't seem to benefit from including traces, while GPT-4 gets a tangible lift over the error-only prompts. 
Overall, on two out of three datasets, trace-based prompts significantly improve the ability of GPT-4 to generate working bug fixes. 

While GPT-3.5 lagging behind in terms of absolute scores irrespective of prompt type is expected,  more broadly, its inability to benefit from execution traces (even degraded performance) could highlight a qualitative generational gap when it comes to emergent abilities of LLMs. 
Notwithstanding, there remain a few unexplained results, such as the lack of performance gain from using traces on the Refactory dataset and the unusually strong performance of the Self-Debug baseline in one particular configuration. To gain a fine-grained understanding, in the next research question, we focus on studying the varying complexity of execution traces and how they affect downstream APR performance.

\begin{center}
\begin{tcolorbox}[colback=gray!10,
                  colframe=black,
                  width=7.5cm,
                  arc=1mm, auto outer arc,
                  boxrule=0.9pt,
                 ]
 \textbf{RQ1 Summary.} Trace prompts do not consistently outperform Error Prompts on program repair. 
\end{tcolorbox}
\end{center}

\subsection{RQ2. How does trace complexity affect the effectiveness of trace-based prompts?}
\label{sec:RQ2}

\begin{figure}[!t]
    \centering
\begin{subfigure}[t]{0.45\textwidth}
        \includegraphics[width=\textwidth,trim={0.9cm 0.9cm 0 0},clip]{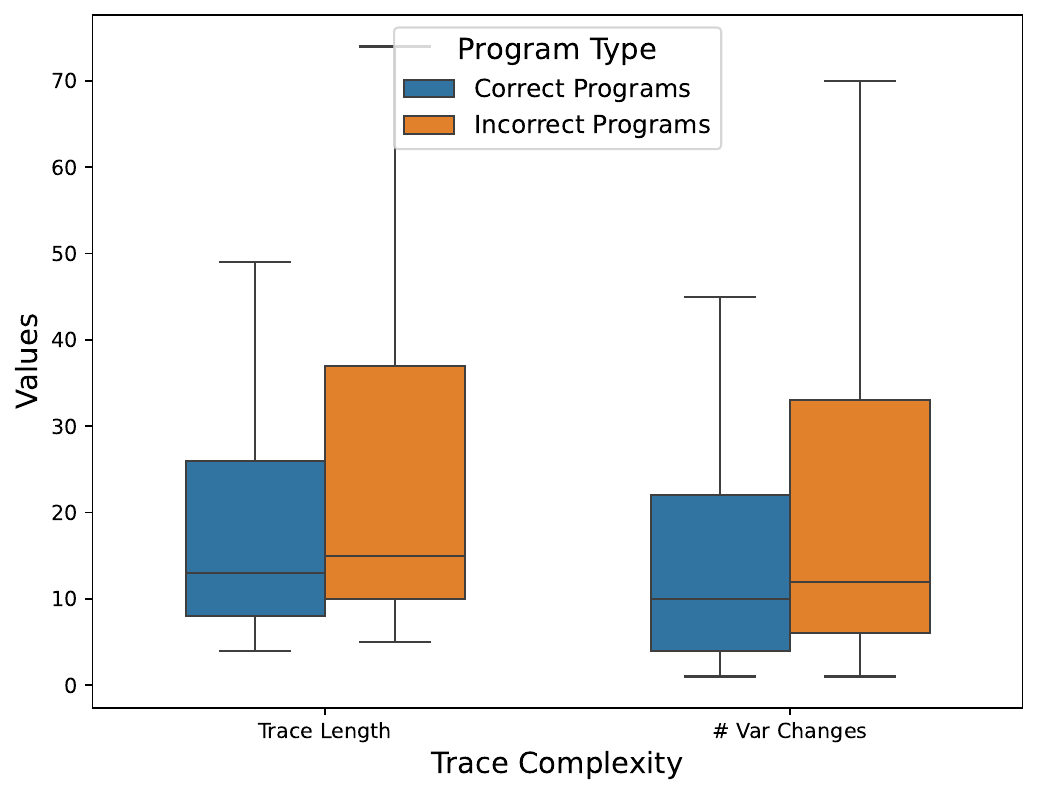}
        \caption{HumanEval-Java Dataset with GPT-4 }
    \end{subfigure}

    \begin{subfigure}[t]{0.45\textwidth}
        \includegraphics[width=\textwidth,trim={0.9cm 0.9cm 0 0},clip]{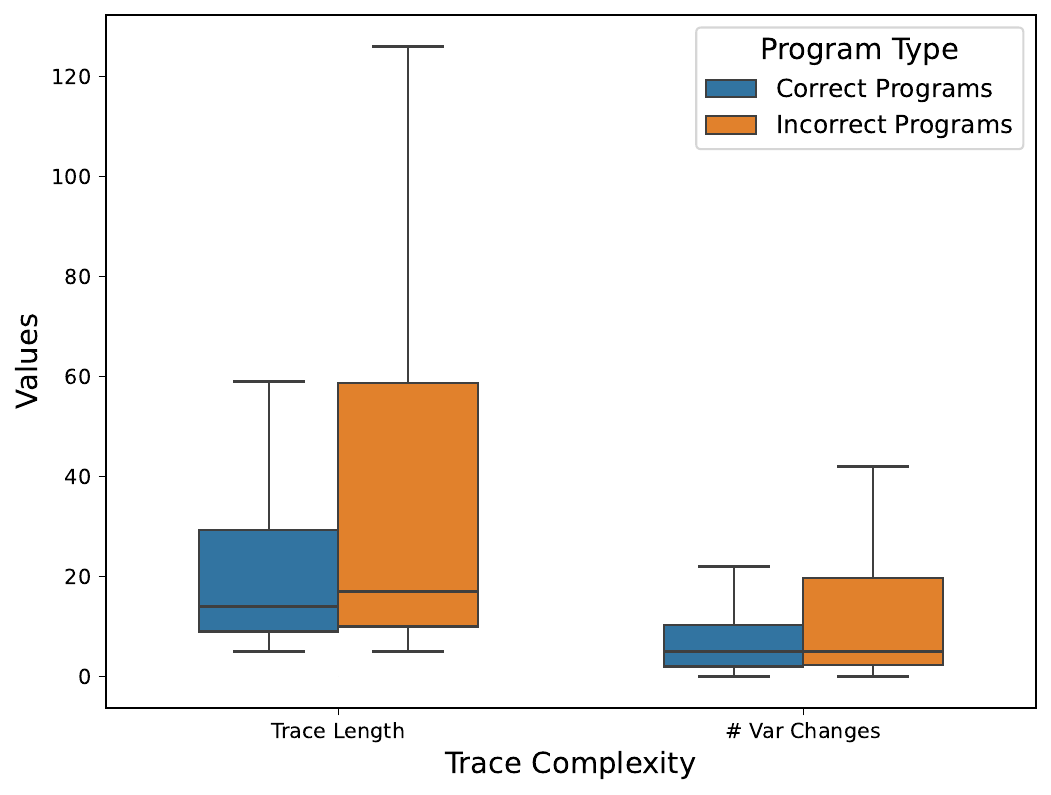}
        \caption{RunBugRun Dataset with GPT-4 }
    \end{subfigure}

    \begin{subfigure}[t]{0.45\textwidth}
        \includegraphics[width=\textwidth,trim={0.9cm 0.9cm 0 0},clip]{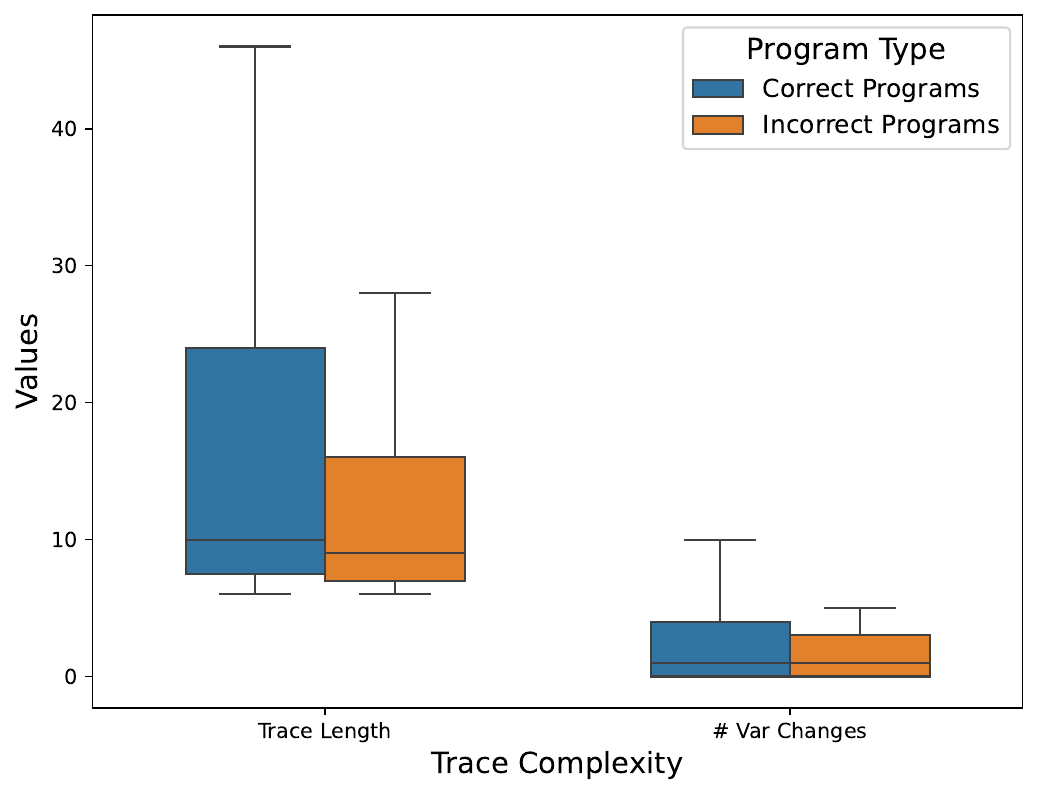}
        \caption{Refactory Dataset with GPT-4 }
    \end{subfigure}
     \caption{Distributions of trace lengths and variable changes across correct vs incorrect program fixes generated by  GPT-4. Analysis for GPT-3.5, showing a similar trend, is given in the appendix.}
    \label{fig:complexity-gpt4}
\end{figure}

Unlike other elements of the prompt, execution traces are dynamic in nature and are highly dependent on a particular input as much as the program itself. Additionally, execution traces can be dramatically different in the presence of a runtime error, compared to when the function finishes executing correctly (even if the returned value itself is wrong).
Thus, variations in trace complexity could be a crucial factor in how beneficial their inclusion is in the prompt. On the one hand, traces that are too short may not provide much information beyond what is already conveyed by the program itself and the failing test case. On the other hand, overly long and complex traces may overwhelm LLMs' long context and ultimately confuse it.
We believe there is a sweet spot at which the inclusion of traces is most beneficial.
As such, we observed great variability with respect to the overall trace length, as well as in the number and type of individual state changes. 

To gain insights into nuanced differences among our evaluated datasets, we compute the statistics of overall trace length and the number of variable modifications in all prompts, while differentiating by whether the resulting fix was correct (Figure \ref{fig:complexity-gpt4}). 
For both the HumanEval-Java and RunBugRun datasets, median\footnote{In all datasets and for both correct and failing fixes we observed the presence of extremely long traces in excess of 10,000 entries. Additionally, a significant number of trace prompts got truncated (5\% for Refactory and almost 10\% for RunBugRun). 
} trace length and number of variable modifications were significantly higher for failing fixes than for the correct ones. This corroborates our presupposition that longer traces could undermine rather than help APR. 
Conversely, in the Refactory dataset, somewhat contrary to our intuitions, for failing fixes median trace length were actually lower than for successful fixes. Regarding the number of variable modifications, the median was just one, compared to 5 in RunBugRun. This disparity implies variable modifications could play a key role in the effectiveness of a trace for APR.






\begin{center}
\begin{tcolorbox}[colback=gray!10,
                  colframe=black,
                  width=7.5cm,
                  arc=1mm, auto outer arc,
                  boxrule=0.9pt,
                 ]
 \textbf{RQ2 Summary.} The prompts having longer execution traces have a lower chance of generating a correct fix. 
\end{tcolorbox}
\end{center}

\section{Impact of Modified Traces}
\label{sec:RQ3}
\label{sec:rq3}

As we find that longer trace length could have a negative impact in the effectiveness of GPT models, we focus on modified trace strategies and their impact on the effectiveness of the model. In this section, we discuss one research question.

\noindent\textbf{RQ3.} Can the format of traces be optimized to guarantee gains for APR?


\subsection{Modified Traces}

\noindent\textbf{Collated Execution Traces.}
Even though execution traces for both languages reference code lines from the original program, they are placed in the standalone section of the prompt, separate from the program itself.
In order to thoroughly ablate trace format, we experimented with combining the two by placing each trace entry directly to its corresponding line of code as an inline comment \footnote{In case of multiple passes through the same line e.g., variable changes within a loop, we concatenate each of the traced events by a new line, providing a full history of state changes at that line.}. The rationale behind this design choice is to consolidate the two types of information in the common location, potentially freeing the LLM from having to constantly cross-reference between them. 

\noindent\textbf{LLM-Optimized Execution Traces (OPT).}
While in a general case deterministic traces provide valuable information regarding variable changes, logging every single event is not always ideal.
In scenarios such as infinite loops, traces end up repeating the same information, while also unboundedly growing in length. 
It can thus be desirable to optimize potentially lengthy traces by condensing superfluous information.
To optimize execution traces, we prompt a long context GPT4-32k model with the deterministic execution trace and an instruction to generate a shorter version of it, optimized for downstream APR.

\noindent\textbf{Confidence Based Prompt Selection (Conf OPT).}
In addition to modifying the format or content of the prompt itself, we experimented with a simple prompt routing mechanism based on pre-querying LLM's confidence about correctly solving a program repair task using the deterministic trace. If the confidence level is low, we fallback onto using an LLM-optimized trace instead.

\begin{figure}
  \begin{center}
    \includegraphics[width=0.5\textwidth]{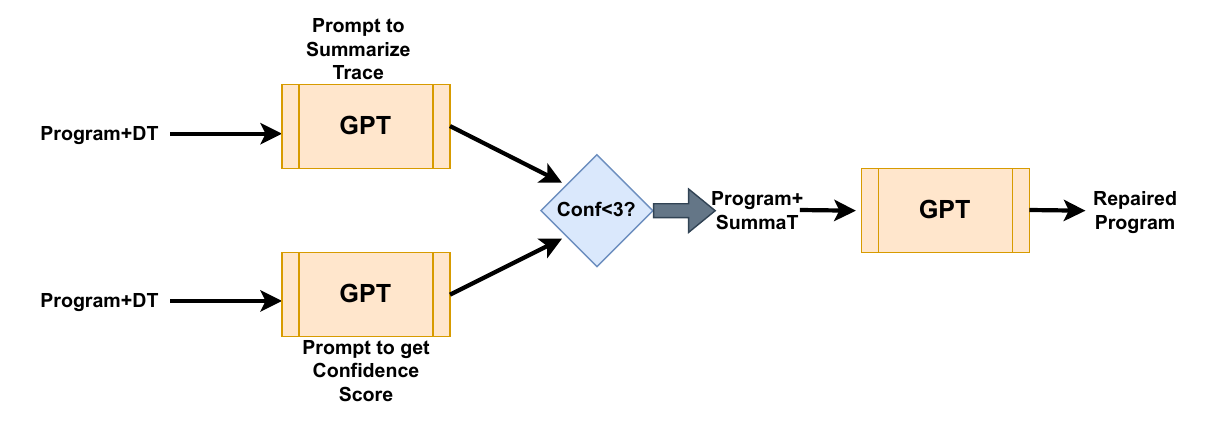}
  \end{center}
  \caption{The Flow of Conditional Selection of Traces}
  \label{fig:cond}
\end{figure}

We have considered multiple ways to find the confidence value of the model. 
One possible way is to prompt the model to find whether it's confident or not (boolean) to use a specific prompt to repair a program. But on a small prompt set, we find that the model outputs that it is always confident for all inputs. 
Additionally, another way is to feed both prompts and ask the model for which prompt it is more confident to repair the program. However, based on the findings of recent work~\cite{huang2023embrace}, LLM might be biased for a specific position (prompt one or prompt two). Hence, based on the findings of Huang et al.~\cite{huang2023embrace}, we use a Likert-scale based confidence score. Given a score range of 1-5, if the confidence score provided by the model is less than 3, we consider that the model has low-confidence. The approach is shown in Figure \ref{fig:cond}.

\noindent\textbf{Trace-length Based Prompt Selection (TRL OPT).}
As we have found through investigating RQ2 that trace prompts work well if the trace length is within a specific range; hence, switching to a different prompt given a longer trace might be beneficial. In this technique, instead of using a confidence score like Conf OPT, we use trace length for routing between prompts. The routing is investigated in two settings: trace prompt and OPT prompt, and trace prompt and error prompt. If the trace length is less than $N$, we use trace prompt, or we use OPT prompt or error prompt based on the setting. We use the following $N$ values for the experiment: 25,30,35,40,45,50.

\begin{table*}[t]
  \centering
  \resizebox{0.9\textwidth}{!}{
    \begin{tabular}{l|l|c|c|c|c|c|c}
    \toprule
    \multirow{2}{*}{\textbf{Metric}} & \multirow{2}{*}{\textbf{Method}} & \multicolumn{3}{c|}{\textbf{GPT-3.5}} & \multicolumn{3}{c}{\textbf{GPT-4}} \\
    \cline{3-8}
     &  & \textbf{Refactory} & \textbf{HumanEval-Java} & \textbf{RunBugRun} & \textbf{Refactory} & \textbf{HumanEval-Java} & \textbf{RunBugRun} \\
    \midrule
    \multirow{4}{*}{\textbf{CFA}}     
    & \textbf{Collated Trace} & 0.452 & 0.391 & 0.381 & 0.656 & 0.531 & 0.483 \\
    & \textbf{OPT Trace} & 0.502 & \textbf{0.430} & \textbf{0.472} & 0.753 & \textbf{0.572} & 0.570 \\
    & \textbf{Conf OPT Trace} & 0.368 & 0.380 & 0.429 & 0.735 & 0.549 & 0.527 \\
    & \textbf{TRL OPT Trace (EP)} & 0.490 & 0.312 & 0.457 & 0.742 & 0.492 & 0.549 \\
    & \textbf{TRL OPT Trace (OPT)} & 0.493 & 0.353 & 0.466 & 0.737 & 0.473 & 0.574 \\
    & \textbf{RQ1 Best} & \textbf{0.525} & 0.380 & 0.465 & \textbf{0.791} & 0.511 & \textbf{0.602} \\
    \midrule
    \multirow{4}{*}{\textbf{CPA}}     
    & \textbf{Collated Trace} & 0.587 & 0.497 & 0.407 & 0.818 & 0.681 & 0.508 \\
    & \textbf{OPT Trace} & 0.601 & 0.535 & \textbf{0.497} & 0.862 & 0.713 & 0.589 \\
    & \textbf{Conf OPT Trace} & 0.384 & 0.522 & 0.453 & 0.847 & \textbf{0.732} & 0.550 \\
    & \textbf{TRL OPT Trace (EP)} & 0.623 & 0.528 & 0.484 & 0.826 & 0.694 & 0.589 \\
    & \textbf{TRL OPT Trace (OPT)} & 0.623 & \textbf{0.573} & 0.491 & 0.826 & 0.675 & 0.603 \\
    & \textbf{RQ1 Best} & \textbf{0.659} & 0.547 & 0.484 & \textbf{0.884} & 0.713 & \textbf{0.629} \\
    \bottomrule
    \end{tabular}%
    }
    \caption{\textbf{RQ3 Quantitative Results.} CFA = Correct Fix Accuracy, and CPA = Correct Program Accuracy.}
  \label{tab:Per_c10_2_2_transposed}
\end{table*}
\subsection{RQ3 Results.}


The results could be found in Table \ref{tab:Per_c10_2_2_transposed}. 
For ease of comparison, for each dataset and model we include the best performing strategy from RQ1, which could be either error prompt, trace prompt or the Self-Debug baseline.
Of the three trace modification strategies, LLM-Optimized prompts (OPT) provide the most consistent performance gains on both CFA and CPA metrics. 
With respect to CPA, for all dataset and model pairs, OPT is among the top three performing prompting techniques. The CFA values for OPT are even more commendable, whereas, for three out of six model-dataset pairs, OPT has the best CFA (second best in the other three). Furthermore, this confirms our implication from RQ2 that less complicated traces are better for prompting for program repair tasks. 

For confidence-based prompt selection, while we find the CFA and CPA values are comparatively better for GPT4, the performance in GPT3.5 is worse. This would imply that GPT-4 is significantly better in providing confidence scores for prompts than GPT-3.5. But, as the performance is significantly worse than OPT on average, the application of the method for GPT-4 is still not reliable.

For trace length-based prompt selection, we only report the best results in the table.
We have two findings here; first, although routing could improve the CFA and CPA values more than individual prompts, we find that only for the GPT-3.5 model and HumanEvalJava dataset could routing get the best CPA score among all considered techniques. Second, changing the value of $N$ would have a limited impact on CFA and CPA values. Overall, we could not find any strong result suggesting that routing between techniques based on trace length might be significantly beneficial. Detailed results could be found in Figures \ref{fig:dsc21} and \ref{fig:dsc22} (in Appendix).

Lastly, collated trace prompts disappointingly do not provide an improvement over trace prompts. One possible explanation is a lack of exposure to this format during LLM training as code doesn't normally include inline comments about state changes.
Second, inline traces within loops can ``stretch'' the length of the program quite a bit, possibly diluting LLMs attention to the continuation of the program after the loop. In our probing studies of LLM trace understanding, we find that, indeed, LLMs struggle to keep up with variable changes across multiple iterations. Finally, the problem of truncation becomes more severe with collated traces, as not just the trace but also part of the original problem could be excluded from the prompt.

\begin{center}
\begin{tcolorbox}[colback=gray!10,
                  colframe=black,
                  width=7.5cm,
                  arc=1mm, auto outer arc,
                  boxrule=0.9pt,
                 ]
 \textbf{RQ3 Summary.} Optimized trace prompt is the most consistent type of prompting technique, specifically for CFA metric.
\end{tcolorbox}
\end{center}

\section{Additional Studies}
\label{sec:disc}
\subsection{Trace-based prompting compared to finetuning a smaller model.}

In this RQ, we focus on evaluating if fine-tuning a small-sized LLM would generate better results w.r.t program repair rather than prompting GPT models with different prompts. For that purpose, we fine-tune the deepseek-coder-1.3b-instruct~\footnote{https://github.com/deepseek-ai/DeepSeek-Coder} model with training data extracted from HumanEval-Java and RunBugRun datasets. Finally, we compare the program repair performance of fine-tuning and prompting-based techniques on test data.

\noindent\textbf{Finetuning Setup.}
 Our finetuning approach is inspired by TraceFixer~\cite{bouzenia2023tracefixer}, which finetunes a CodeT5 model using the buggy program's code, its execution trace, and the desired state of the program. As we didn't have access to TraceFixer's code, we implemented our own finetuning pipeline.
In our case, the input to the model consists of a buggy program, the failing test case results, and corresponding execution traces. During training, the correct version of the program is included in the prompt, while during inference it is omitted, to be filled in by the model. For each dataset, 80\% of the problems are randomly selected for training, and the rest are reserved for testing. This accounts for 459 samples for RunBugRun and 517 samples for HumanEval-Java datasets.
We use the training settings and parameters suggested by deepseek-coder developers to finetune this model. Details of these parameters can be found in the model's repo.

\begin{figure}[t]%
	\centering
	\includegraphics[width=0.48\textwidth]{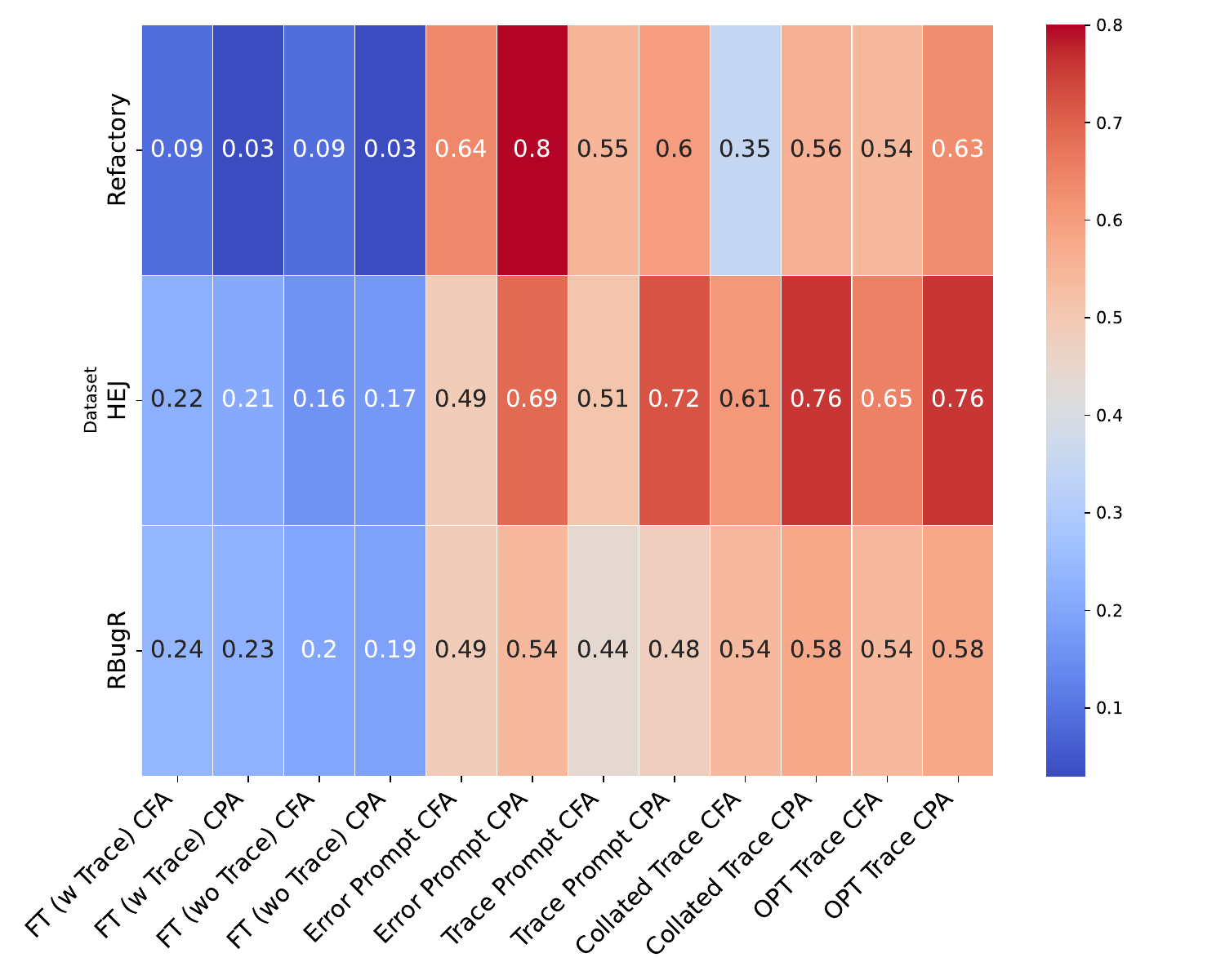}
    \caption{CPA and CFA of Prompting-Based Techniques vs. Model Fine-Tuned for APR Tasks. }
	\label{fig:ft}
\end{figure}

\noindent\textbf{Result.} Figure \ref{fig:ft} shows the results. For comparison purposes, we calculate CPA and CFA for prompting-based techniques on the same test programs. It can be noted that all the prompting techniques outperform fine-tuned model's CPA and CFA. It is observed that models fine-tuned with and without trace show lower CPA and CFA than prompting-based techniques. One of the reasons behind the results might be the limited training data for each task. Also, the TraceFixer technique showed better results in the original work, but the number of training examples for TraceFixer was significantly higher, too. In our future work, we plan to use a larger training dataset and larger models for finetuning.

\subsection{Probing Studies of Trace Understanding}
To gain insights into the observed lack of improvement in APR performance using collated traces and results from using traces for APR in general, we investigate two additional questions via small-scale probing experiments.

\paragraph{Can the LLM align the program with its execution trace?} 
We directly measure the LLM's ability to perform trace collating given a standalone program and its execution trace. The rationale behind this experiment is that if an LLM can do this task with high accuracy, then there is no added benefit of adding collating traces into prompts.
\paragraph{Can the LLM infer the execution trace from the program alone?}
Although the Self-Debug approach implicitly traces through the program's execution, it is never formally evaluated. If an LLM can accurately generate a program's execution trace, then adding such a trace into the prompt would understandably not provide additional value for APR.

For both of these experiments we used GPT-4 on a subset of programs from the Refactory dataset. Since trace prediction behavior can be different depending on whether a function executes successfully or raises an error, for each experiment we differentiate between traces produced for working and failing programs. In addition, due to a limited number of distinct problems in Refactory, we additionally evaluate on the Geeks-for-geeks dataset\footnote{\url{https://github.com/facebookresearch/TransCoder}}. To evaluate the LLM's output, we compute a diff against the ground truth trace or collated trace/program and report the exact match rate, after light post-processing, in Table \ref{tab:trace-understanding} of Appendix \ref{sec:probing-qualitative}. 

Based on these results, trace collating accuracy reaches 88\% on reference Refactory programs, however it degrades by nearly ten percent on programs containing failures. Furthermore, on the more diverse Geeks for geeks dataset, which also eliminates the possibility of prompt leakage, collating performance sharply decreases to just 45\%.

Prediction of a program's execution trace by an LLM from scratch is a significantly more challenging task compared to merely modifying the format of the trace. As a result, the rate of zero-diff trace predictions does not exceed 50\% in the case of reference Refactory programs and is further halved for programs containing failures. Across the Geeks for geeks dataset, only 15\% of generated traces perfectly match the ground truth.
We provide qualitative analysis of a manually reviewed sample of diffs in the appendix.

Despite the impressive ability of GPT-4 at manipulating execution traces neither of the two tasks appear to be trivially solvable. Hence, we conclude real execution traces can contribute information for downstream tasks not yet easily inferrable by strong LLMs such as GPT-4.


\section{Conclusion}

In this study, we examined the impact of incorporating program execution traces into prompts on the program repair capabilities of the GPT model family. Our findings indicate that trace-based prompts do not consistently outperform error-based prompts; their effectiveness varies with the dataset and LLM used. Analysis reveals that longer traces and more variable assignments reduce prompt effectiveness. Using this insight, we developed variations of trace-based prompts, finding that LLM-optimized traces offer more consistent improvements without limiting trace complexity heuristically. We validated our results against a fine-tuned baseline and found that LLMs have limited capacity for trace generation, explaining the weaker performance of the Self-Debug baseline and highlighting the potential utility of traces in code tasks.

\section{Disclaimer}
Disclaimer: This paper was prepared for informational purposes   by the Artificial Intelligence Research group of JPMorgan Chase \& Co. and its affiliates ("JP Morgan'') and is not a product of the Research Department of JP Morgan. JP Morgan makes no representation and warranty whatsoever and disclaims all liability, for the completeness, accuracy or reliability of the information contained herein. This document is not intended as investment research or investment advice, or a recommendation, offer or solicitation for the purchase or sale of any security, financial instrument, financial product or service, or to be used in any way for evaluating the merits of participating in any transaction, and shall not constitute a solicitation under any jurisdiction or to any person, if such solicitation under such jurisdiction or to such person would be unlawful.


\bibliography{acmart}
\onecolumn
\appendix

%


\section{Rationale behind prompt choice and other prompts considered.}
\label{sec:prompt-rationale}
In our preliminary experiment error-based prompts always performed better than program-only prompts. 
Hence, we use Error Prompts as a point of comparison for Trace Prompts, foregoing prompts only containing the buggy program.
Furthermore, we explored the option of including all failing test cases in the same prompt, however that did not provide a lift compared to a single test case, and overall performed slightly worse. We hypothesize multiple test cases could be more helpful for program generation to help define the space of valid solutions, whereas in APR the buggy function itself provides a bulk of information for fixing a bug, and a single failing test case, while inexhaustive, is generally sufficient for setting the LLM on the right path to finding a fix.
The use of few-shot prompts, while feasible for improving the accuracy of error-based prompts, is problematic for traces as it can greatly increase the overall length of the prompt, potentially exceeding the 8k context window.

\begin{figure}[!t]
    \centering
    \begin{subfigure}[b]{0.45\textwidth}
        \includegraphics[width=\textwidth,trim={0.9cm 0.9cm 0 0},clip]{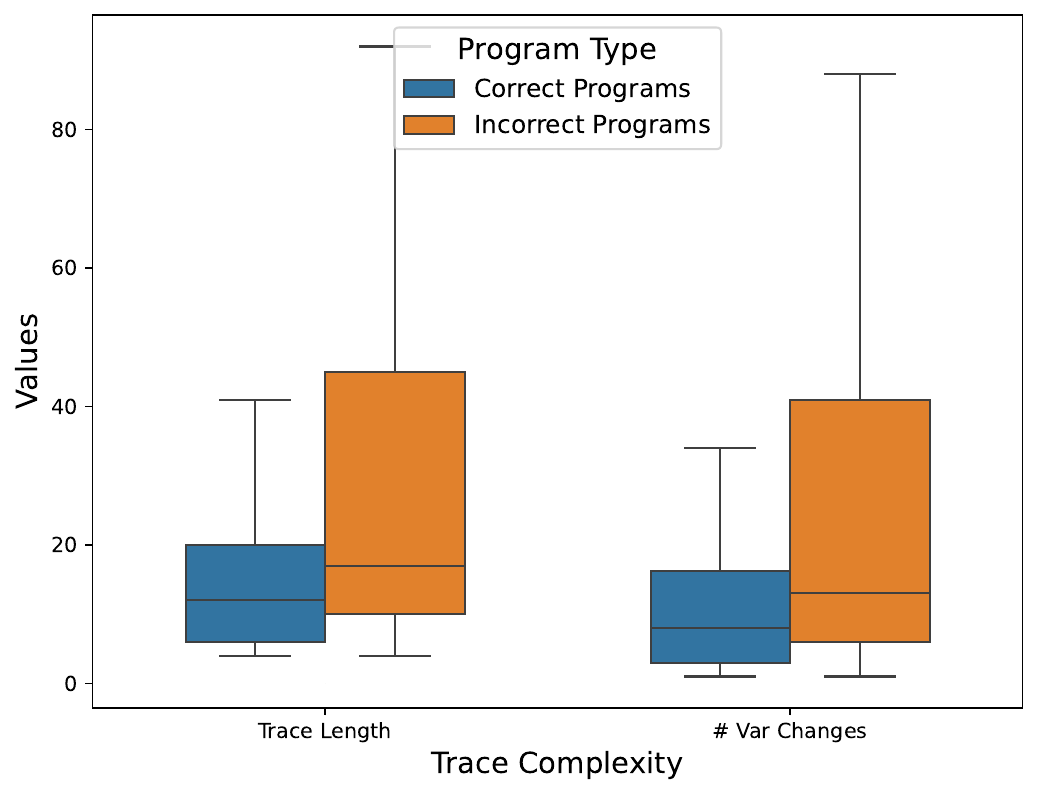}
        \caption{HumanEval-Java Dataset with GPT-3.5}
        \label{fig:HEJ_GPT35}
    \end{subfigure}

    \begin{subfigure}[b]{0.45\textwidth}
        \includegraphics[width=\textwidth,trim={0.9cm 0.9cm 0 0},clip]{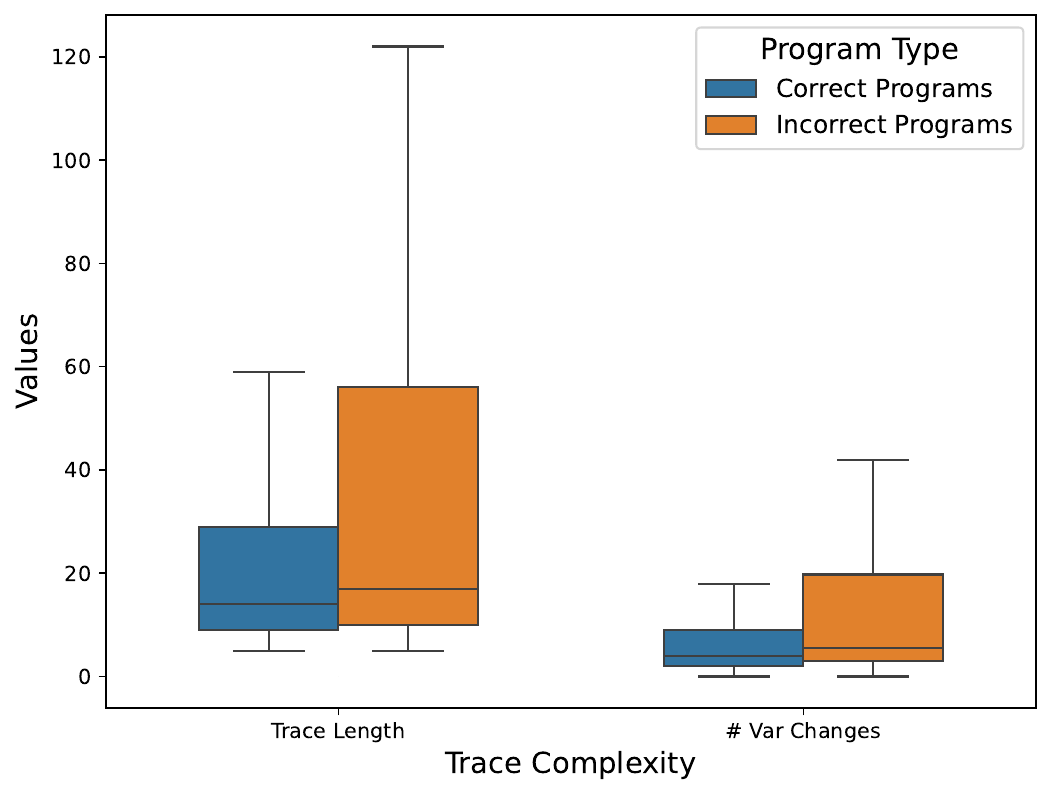}
        \caption{RunBugRun Dataset with GPT-3.5}
        \label{fig:RBR_GPT35}
    \end{subfigure}

    \begin{subfigure}[b]{0.45\textwidth}
        \includegraphics[width=\textwidth,trim={0.9cm 0.9cm 0 0},clip]{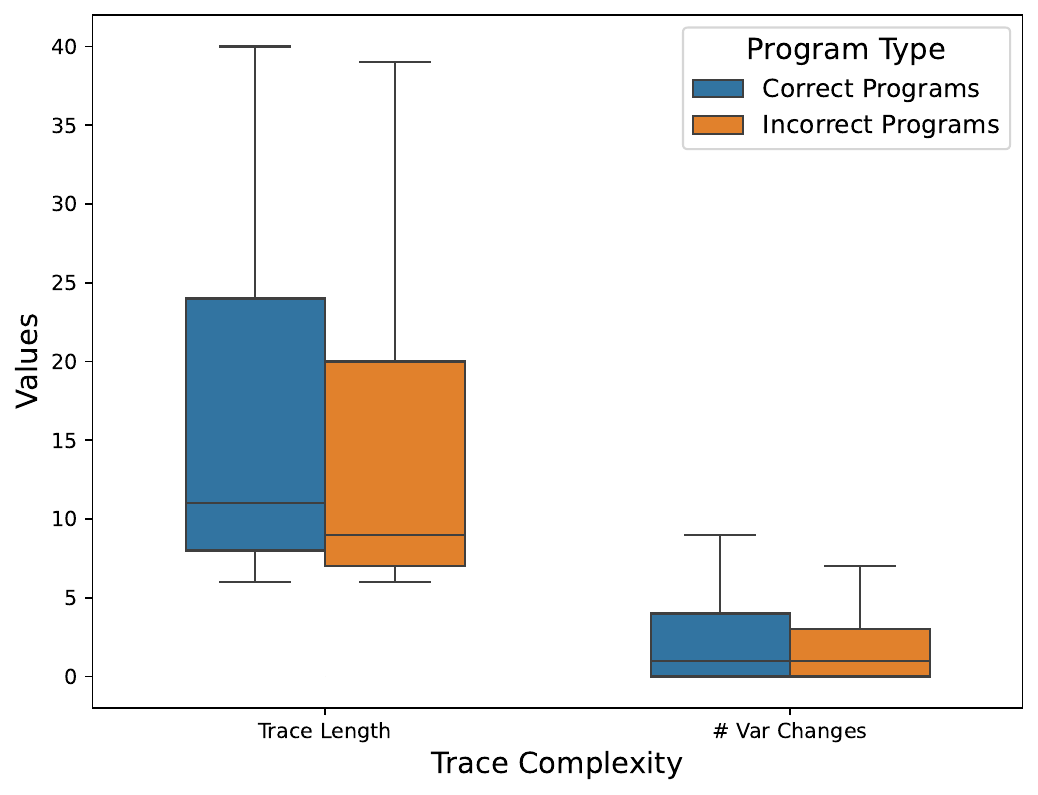}
        \caption{Refactory Dataset with GPT-3.5}
        \label{fig:REF_GPT35}
    \end{subfigure}
     \caption{Distributions of trace lengths and variable changes across correct vs incorrect program fixes generated by GPT-3.5}
    \label{fig:tranly}
\end{figure}

\begin{figure}[!t]
    \centering
    \begin{subfigure}[b]{0.75\textwidth}
        \includegraphics[width=\textwidth]{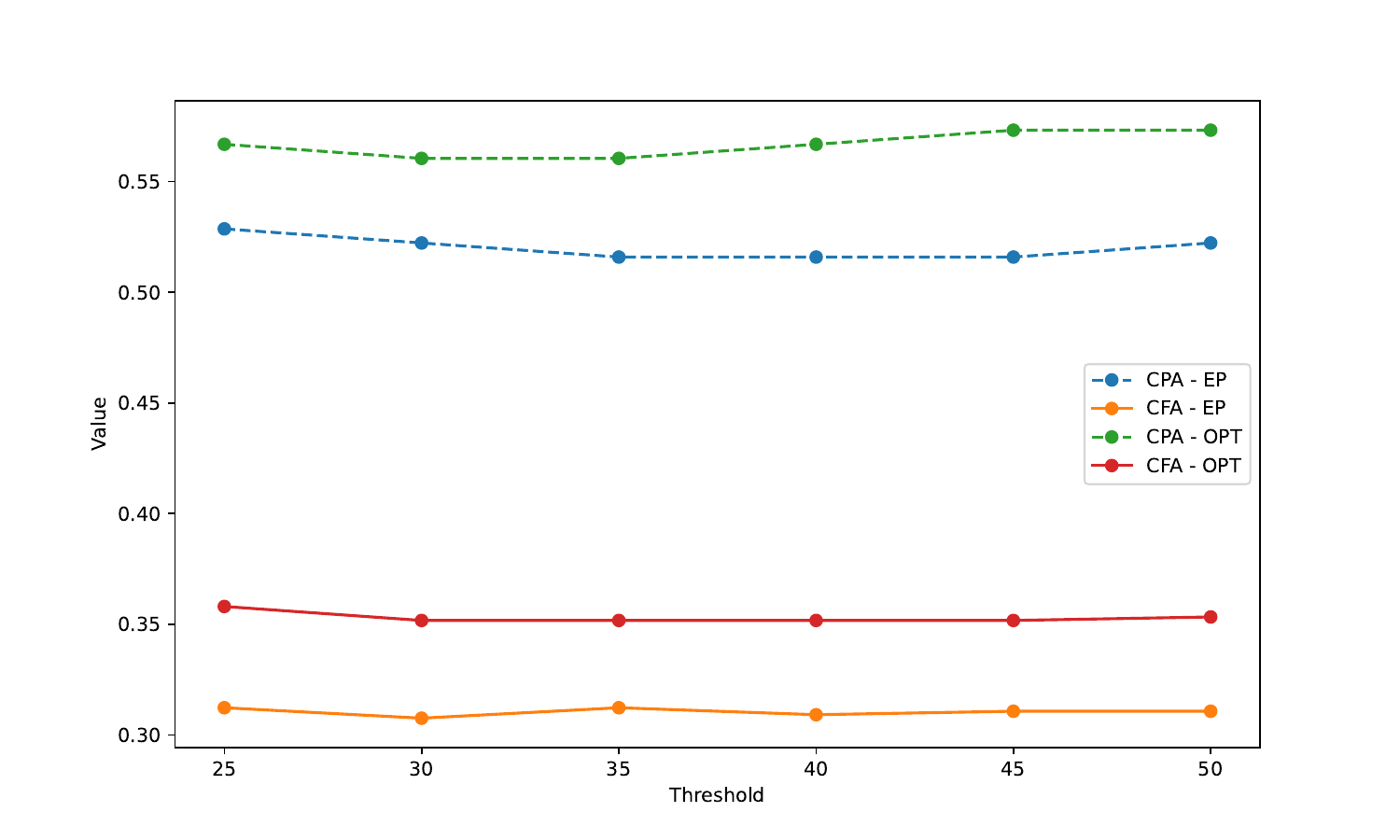}
        \caption{HumanEval-Java Dataset with GPT-3.5}
        \label{fig:HEJ_GPT35_dsc}

    \end{subfigure}

    \begin{subfigure}[b]{0.75\textwidth}
        \includegraphics[width=\textwidth]{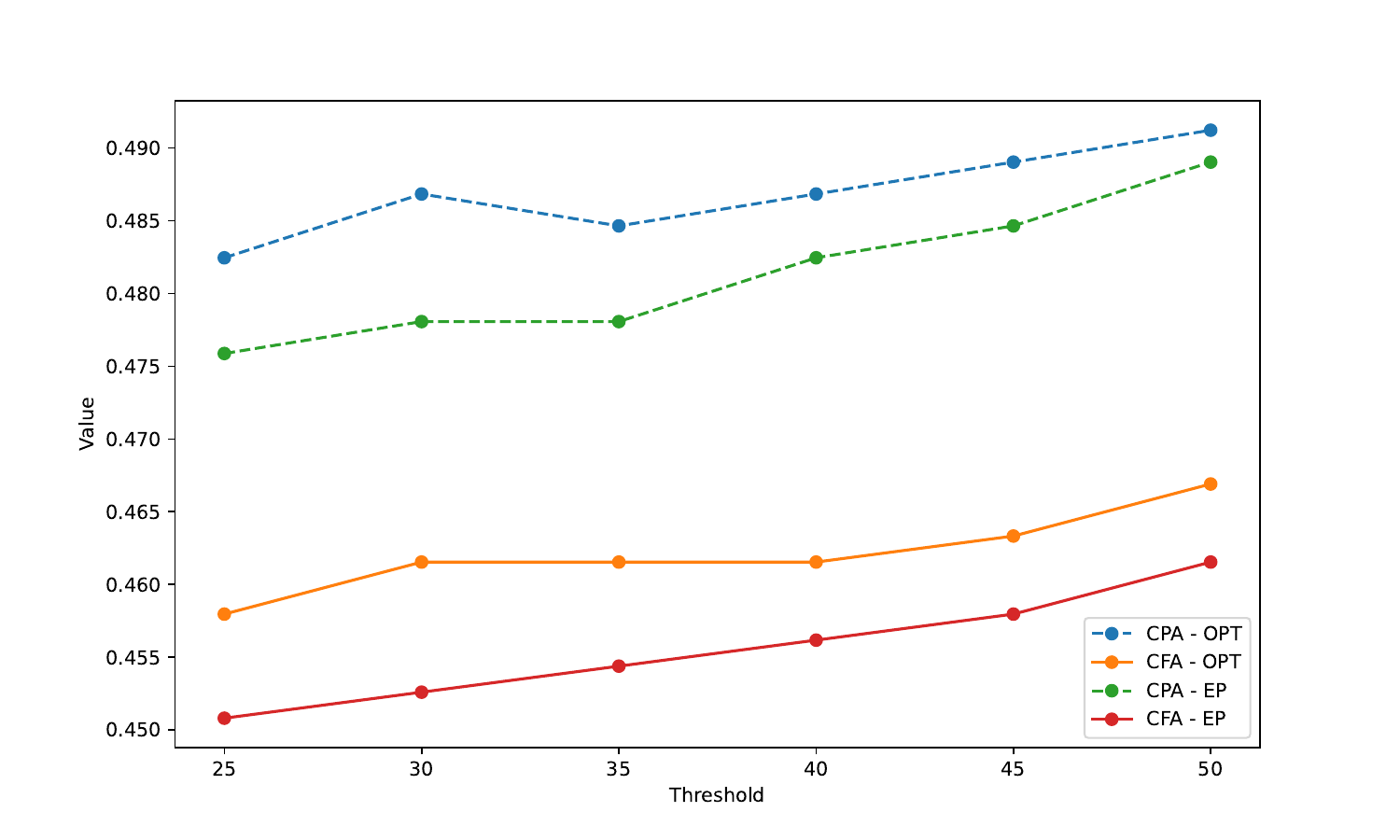}
        \caption{RunBugRun Dataset with GPT-3.5}
        \label{fig:RBR_GPT35_dsc}

    \end{subfigure}

    \begin{subfigure}[b]{0.75\textwidth}
        \includegraphics[width=\textwidth]{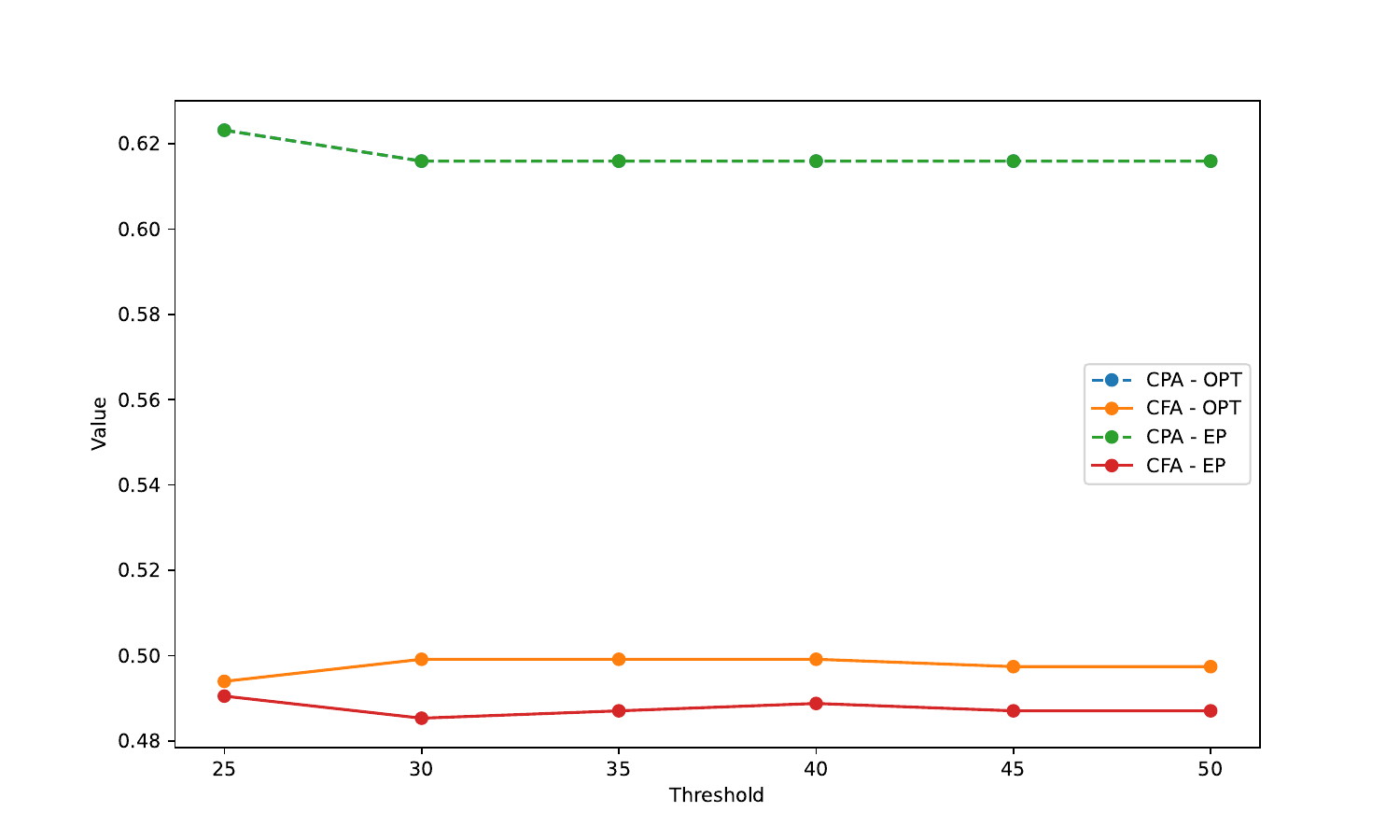}
        \caption{Refactory Dataset with GPT-3.5}
        \label{fig:REF_GPT35_dsc}

    \end{subfigure}
     \caption{Ablations of trace length threshold values used with the routing strategy for GPT3.5}
    \label{fig:dsc21}
\end{figure}

\begin{figure}[!t]
    \centering
    \begin{subfigure}[b]{0.75\textwidth}

         \includegraphics[width=\textwidth]{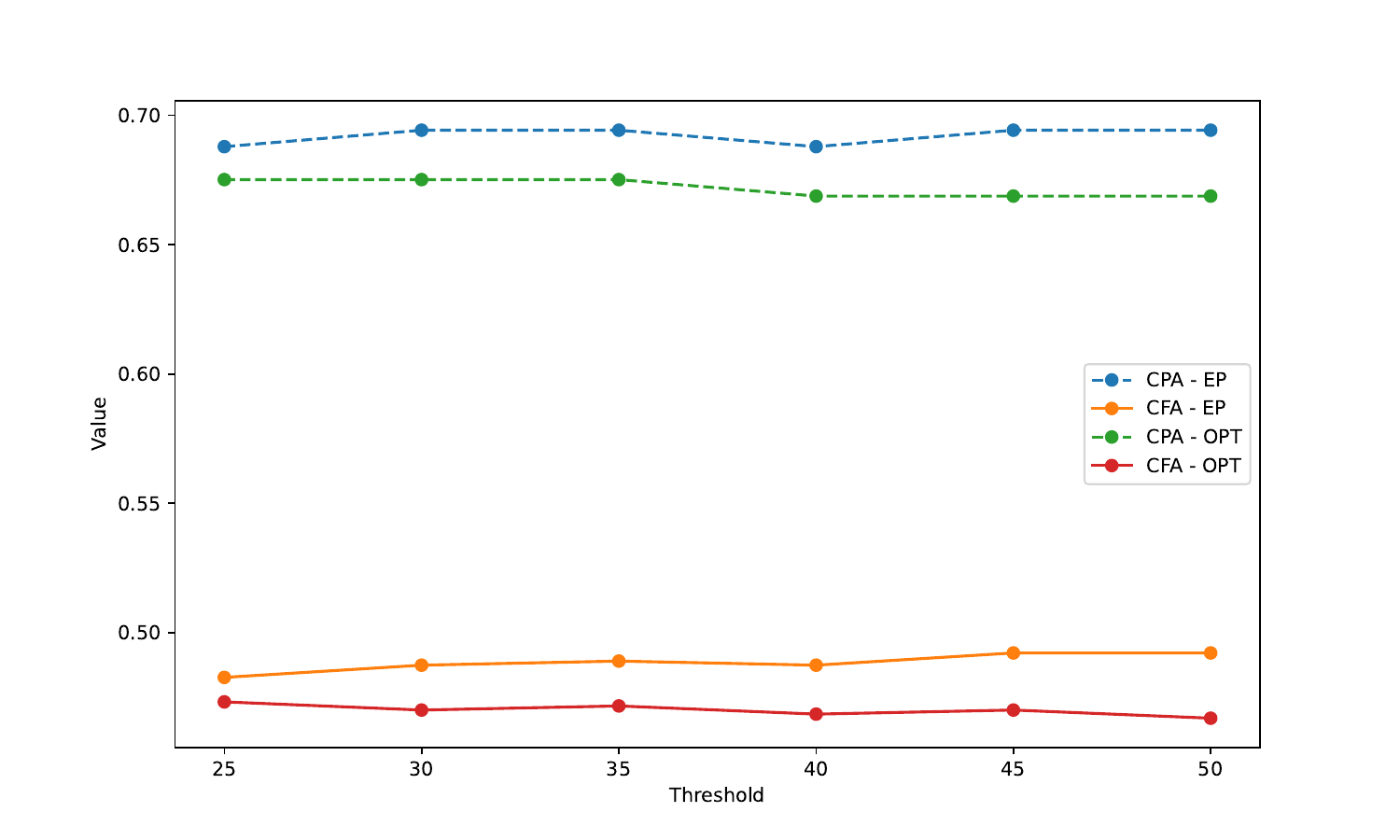}
        \caption{HumanEval-Java Dataset with GPT-4}
        \label{fig:HEJ_GPT4_dsc}
    \end{subfigure}

    \begin{subfigure}[b]{0.75\textwidth}

        \includegraphics[width=\textwidth]{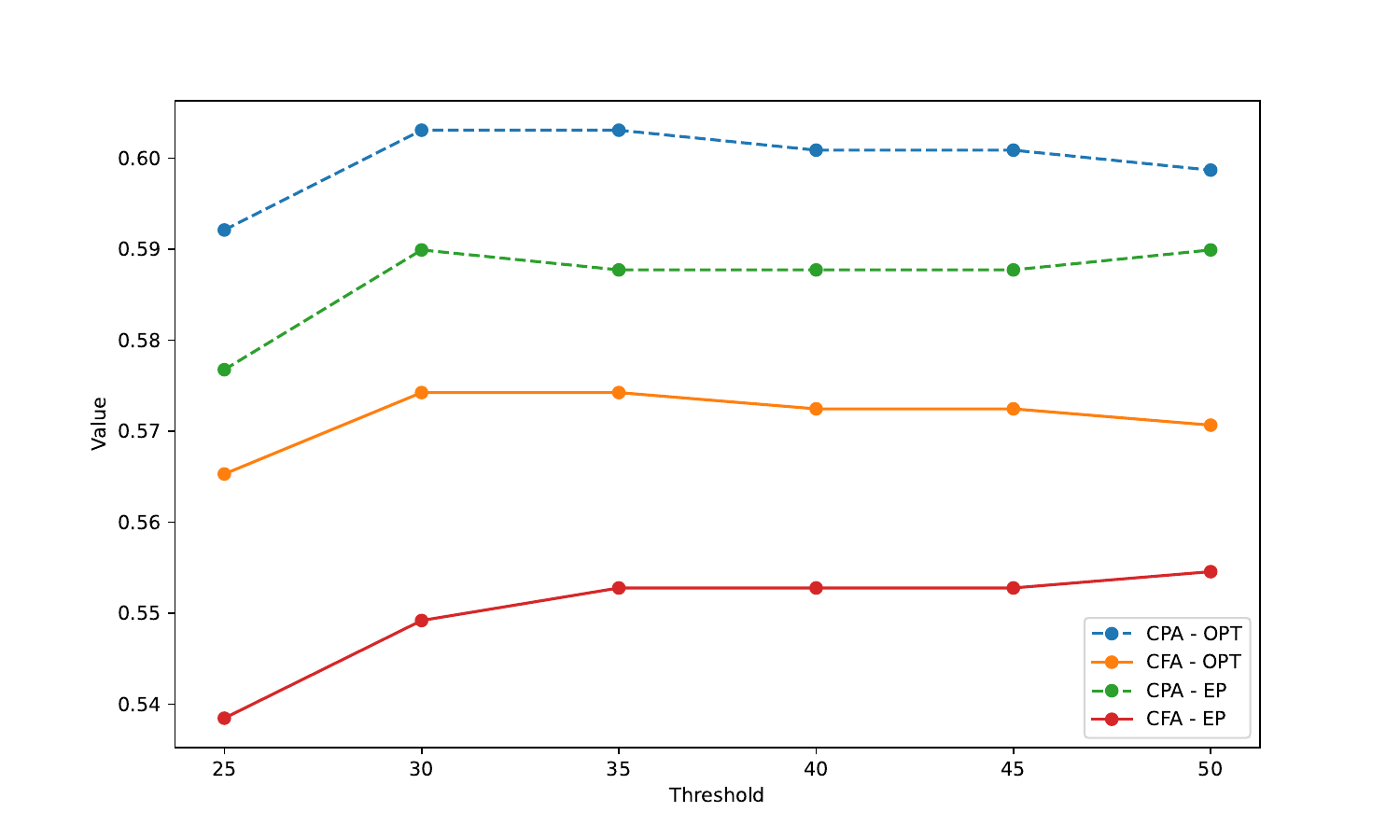}
        \caption{RunBugRun Dataset with GPT-4}
        \label{fig:RBR_GPT4_dsc}
    \end{subfigure}

    \begin{subfigure}[b]{0.75\textwidth}

        \includegraphics[width=\textwidth]{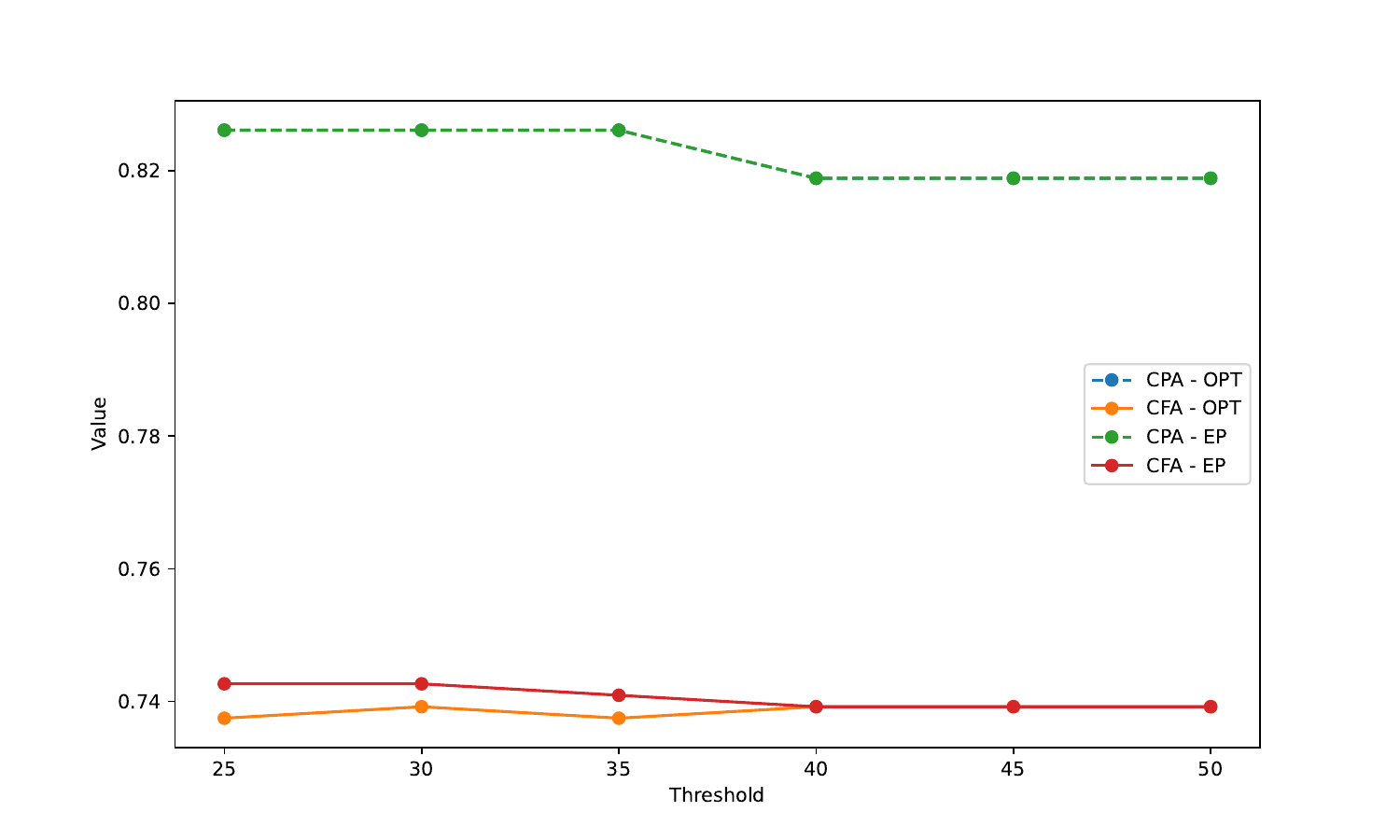}
        \caption{Refactory Dataset with GPT-4}
        \label{fig:REF_GPT4_dsc}
    \end{subfigure}
    
     \caption{Ablations of trace length threshold values used with the routing strategy for GPT-4.}
    \label{fig:dsc22}
\end{figure}
\clearpage

\subsection{Trace Understanding Probing Studies Full Results and Qualitative Findings}
\label{sec:probing-qualitative}
\begin{table}[h!]
\centering
\caption{Trace Understanding Probing Results}
\begin{tabular}{lllllll}
\cline{1-4}
\multicolumn{1}{|l|}{Dataset partition (\#prompts)}                 & \multicolumn{1}{l|}{Trace Collating} & \multicolumn{2}{l|}{Trace Prediction} &  &  &  \\ \cline{1-4}
\multicolumn{1}{|l|}{Refactory reference (34)} & \multicolumn{1}{l|}{88\%}            & \multicolumn{2}{l|}{50\%}             &  &  &  \\ \cline{1-4}
\multicolumn{1}{|l|}{Refactory fail (38)} & \multicolumn{1}{l|}{79\%}            & \multicolumn{2}{l|}{26\%}             &  &  &  \\ \cline{1-4}
\multicolumn{1}{|l|}{Geeks for geeks (300)} & \multicolumn{1}{l|}{45\%}            & \multicolumn{2}{l|}{15\%}             &  &  &  \\ \cline{1-4}
\label{tab:trace-understanding}
\end{tabular}
\end{table}
We manually reviewed a sample of diffs to gain qualitative insights of LLM trace manipulation behavior. 
Most discrepancies between ground truth and either LLM-collated or predicted traces are due to additions or deletions of variable modifications from the trace. In particular, within loops, the LLM tends to either miss or add extra variable modifications., which could hit at a potential limitation in the depth of reasoning and memory.
In the task of trace prediction from scratch, the second most erratic behavior is around predicting function returns, which can amount to both wrong value and wrong placement within the execution flow. Interestingly, in addition to generating traces, the LLM consistently attempts to fix code formatting, and in many cases optimizes away code branches not taken. Similarly, in the presence of execution failures, the LLM is unreliable at correctly predicting exceptions -- either predicting exception types not commonly raised by a given operation, missing the exception altogether or, in some cases, patching the code to prevent an exception. Miscellaneous observed other discrepancies are due to the LLM adding superfluous commentary, trace formatting mistakes and hallucination of object hashes and other literals.
\clearpage


\end{document}